\documentclass[10pt,twocolumn,letterpaper]{article}

\usepackage[pagenumbers]{cvpr} 

\usepackage{graphicx}
\usepackage{amsmath}
\usepackage{amssymb}
\usepackage{booktabs}
\usepackage[american]{babel}
\usepackage{cite}
\usepackage{overpic}
\usepackage{caption}
\usepackage{makecell}
\usepackage{multirow}
\usepackage[table,xcdraw]{xcolor}
\usepackage{lipsum}

\usepackage{bbding} 

\let\OLDthebibliography\thebibliography
\renewcommand\thebibliography[1]{
    \OLDthebibliography{#1}
    \setlength{\parskip}{0pt plus 0.5ex}
    \setlength{\itemsep}{0pt plus 0.5ex}
}

\usepackage[pagebackref,breaklinks,colorlinks]{hyperref}

\newcommand*{\affmark}[1][*]{\textsuperscript{#1}}

\newcommand*{\institute}[1]{#1}

\usepackage[capitalize]{cleveref}
\crefname{section}{Sec.}{Secs.}
\Crefname{section}{Section}{Sections}
\Crefname{table}{Table}{Tables}
\crefname{table}{Tab.}{Tabs.}
\Crefname{equation}{Equation}{Equations}
\crefname{equation}{Eqn.}{Eqns.}

\def\cvprPaperID{3582} 
\def\confName{CVPR}
\def\confYear{2022}

\begin{document}

\title{Adaptive Network Combination for Single-Image Reflection Removal:\\A Domain Generalization Perspective}

\author{%
    Ming Liu\affmark[1,2] \ \ Jianan Pan\affmark[1] \ \ Zifei Yan\affmark[1] \ \ Wangmeng Zuo\affmark[1 (\Envelope)] \ \ Lei Zhang\affmark[2]\\
    \institute{\small \affmark[1]Harbin Institute of Technology, Harbin, China},
    \institute{\small \affmark[2]The Hong Kong Polytechnic University, Hong Kong, China}
}

\maketitle

\begin{abstract}
Recently, multiple synthetic and real-world datasets have been built to facilitate the training of deep single image reflection removal (SIRR) models.
Meanwhile, diverse testing sets are also provided with different types of reflection and scenes.
However, the non-negligible domain gaps between training and testing sets make it difficult to learn deep models generalizing well to testing images.
The diversity of reflections and scenes further makes it a mission impossible to learn a single model being effective to all testing sets and real-world reflections.
In this paper, we tackle these issues by learning SIRR models from a domain generalization perspective.
Particularly, for each source set, a specific SIRR model is trained to serve as a domain expert of relevant reflection types.
For a given reflection-contaminated image, we present a reflection type-aware weighting (RTAW) module to predict expert-wise weights.
RTAW can then be incorporated with adaptive network combination (AdaNEC) for handling different reflection types and scenes, i.e., generalizing to unknown domains.
Two representative AdaNEC methods, i.e., output fusion (OF) and network interpolation (NI), are provided by considering both adaptation levels and efficiency.
For images from one source set, we train RTAW to only predict expert-wise weights of other domain experts for improving generalization ability, while the weights of all experts are predicted and employed during testing.
An in-domain expert (IDE) loss is presented for training RTAW.
Extensive experiments show the appealing performance gain of our AdaNEC on different state-of-the-art SIRR networks.
Source code and pre-trained models will available at \url{https://github.com/csmliu/AdaNEC}.
\end{abstract}

\section{Introduction}
\label{sec:intro}
When capturing images through glass, undesired reflection by the glass surface is one of the key factors causing image quality degradation, and the existence of these reflections also hampers many other vision applications such as image classification and autonomous driving.
In recent years, driven by the development of deep neural networks (DNNs), learning-based methods have achieved continuous improvements on single-image reflection removal (SIRR) tasks~\cite{CEILNet, Zhang, BDN, ERRNet, CoRRN, IBCLN, absorption, RAGNet, RMNet, physical, Location-aware, DMGN, VDESIRR}.

For training these methods, generally a large amount of pair-wise images are required.
To this end, several real-world training sets~\cite{Zhang, ERRNet, IBCLN} have been collected.
However, due to the difficulty and cost to collect real-world reflection and clean image pairs, these datasets contain only a limited number of images.
As a remedy, multiple reflection synthesis methods~\cite{CEILNet, BDN, Zhang, RMNet, physical, absorption} are introduced, which can be used for joint-training with real-world datasets.
Albeit the tremendous efforts to synthesize or render realistic reflection-contaminated images, there is still a large gap between synthetic images and real ones.
Besides, the real-world reflection sets are collected under distinct circumstances, leading to diverse reflection types with varying light conditions, glass types, and photographing parameters.

The non-negligible domain gaps between training and testing sets make the generalization to testing images difficult, and the diversity of reflections and scenes in real-world images further makes it a mission impossible to learn a single model for all testing sets.
As a result, previous reflection removal methods usually have to seek a balanced yet sub-optimal solution on diverse reflection types, remaining some questions to tap the potential of SIRR methods:
i)~\textit{how to adjust the learning policy to train deep models generalizing well to testing images} and
ii)~\textit{how to tune the model adaptively during inference to deal with various reflection types in different testing images or datasets}.

Aiming to learn a model that can generalize to an unseen target domain from one or more different yet related source domains, domain generalization (DG) has been successfully applied in many high-level vision tasks such as classification and semantic segmentation~\cite{DG_survey_Zhou, DG_survey_Wang}.
In this paper, we introduce domain generalization into SIRR to answer the above questions, where the rationality and feasibility are shown via theoretical analysis and statistical evaluation (see \cref{sec:method_analysis} for details).
In particular, for each training set (source domain), we train a specific SIRR model, which serves as the domain expert of relevant reflection types.
Then, for a given testing image (target domain), these models can cooperate to perform reflection removal according to their expertise level on the target reflection type.

In order to evaluate the expertise level, we present a reflection type-aware weighting (RTAW) module, which predicts an expert-wise weight for each model.
During training, we follow the leave-one-domain-out (LODO) strategy~\cite{LODO}, \ie, for images from one source set, RTAW only predicts expert-wise weights of other domain experts, and this source set serves as a pseudo target set for improving generalization ability.
RTAW can then be incorporated with the proposed adaptive network combination (AdaNEC) to handle different reflection types and scenes, \ie, generalizing to unknown domains, where the weights of all experts are predicted and employed.
An in-domain expert (IDE) loss is introduced for training the RTAW module together with the reflection removal losses.

Two representative AdaNEC methods, \ie, output fusion (OF) and network interpolation (NI), are provided by considering both adaptation levels and efficiency.
In the OF manner, the outputs of all experts are fused via the weights predicted by RTAW.
While in the NI manner, we interpolate parameters of the experts, which boosts the performance with minor computation costs.
It is worth noting that images from the same testing set can be regarded as a domain due to the similar capturing circumstance.
Therefore, we can further apply the domain-level AdaNEC, which uses the average RTAW weights to get rid of the interference factors in certain images and achieves a more stable prediction.

To evaluate the proposed method, we take three state-of-the-art methods (\ie, ERRNet~\cite{ERRNet}, IBCLN~\cite{IBCLN}, and RAGNet~\cite{RAGNet}) as the backbone reflection removal networks, and incorporate our proposed AdaNEC method into these backbone models\footnote{Note that ERRNet~\cite{ERRNet}, IBCLN~\cite{IBCLN}, and RAGNet~\cite{RAGNet} are the best-performed SIRR methods with training code publicly available currently.}.
Both quantitative and qualitative experimental results show that the AdaNEC-boosted models perform favorably against their original joint-training counterparts, indicating the effectiveness of our AdaNEC method.

In general, the main contribution of this paper involves:
\begin{itemize}
    \item Based on the formation process of reflections, we propose to learn SIRR models from a domain generalization perspective instead of the joint-training scheme in existing methods.
    \item Given SIRR models as domain experts, a reflection type-aware weighting (RTAW) module is presented to predict the expert-wise weights, which can be incorporated with adaptive network combination (AdaNEC) for generalizing to unseen domains. An in-domain expert (IDE) loss is introduced for training RTAW.
    \item Two representative AdaNEC methods, \ie, output fusion (OF) and network interpolation (NI), are provided considering both adaptation levels and efficiency.
    \item Extensive experimental results on different state-of-the-art SIRR networks show that the proposed AdaNEC can achieve appealing performance gains in comparison with their joint-training counterparts.
\end{itemize}

\section{Related Work}
\label{sec:related_work}

Reflection removal methods generally can be divided into two categories, \ie, single-image reflection removal (SIRR) and multi-image reflection removal (MIRR).
We focus on learning-based SIRR in this paper, some MIRR~\cite{Szeliski_2000_CVPR, Sarel_2004_ECCV, Gai_2012_TPAMI, Li_2013_ICCV, Guo_2014_CVPR, Xue_2015_ToG, Yang_2016_CVPR, Sun_2016_MM, Punnappurath_2019_CVPR, liu2020learning, 2005Removing, lei2021robust, 2004Separating, Wieschollek_2018_ECCV, lei2020polarized} and traditional SIRR~\cite{Li_2014_CVPR, Wan_2016_TIP, Levin_2007_TPAMI, Arvanitopoulos_2017_CVPR, Wan_2018_TIP, Shih_2015_CVPR} methods are listed for reference.
In the following, we first review the learning-based SIRR methods and relevant datasets, then briefly introduce the domain generalization and network interpolation methods.

\subsection{Learning-based SIRR Methods}
\label{sec:related_SIRR}
For removing reflection accurately and producing clear results,
Fan \etal~\cite{CEILNet} designed a gradient loss to constrain the transmission edge maps,
Zhang \etal~\cite{Zhang} further proposed an exclusion loss to minimize the gradient correlation of reflection/transmission layers.
To alleviate the misalignment problem, Wei \etal~\cite{ERRNet} leveraged the highest-level VGG~\cite{VGG} features which are insensitive to misalignment.
Reflection distribution is also considered to facilitate SIRR,
Dong \etal~\cite{Location-aware} predicted a reflection confidence map which is integrated into a composition loss, while Li \etal~\cite{RAGNet} defined an explicit mask loss by leveraging reflection strength.
In this paper, we enhance the SIRR  task from a domain generalization perspective, which is orthogonal to and cooperates well with the above methods.

Various network structures have also been explored in SIRR.
CEILNet presented by Fan \etal~\cite{CEILNet} is a two-stage architecture, where the edge maps are predicted before the estimation of the transmission layer.
The two-stage setting is followed and developed by DMGN~\cite{DMGN}, Dong \etal~\cite{Location-aware}, and RAGNet~\cite{RAGNet}, and they estimate the reflection layer in the prior stage instead.
Zhang \etal~\cite{Zhang} stacked the features extracted by multiple layers of a pre-trained VGG-19~\cite{VGG} model at the beginning of SIRR models, which is retained in a series of subsequent works~\cite{ERRNet, DMGN}.
Yang \etal~\cite{BDN} proposed to predict the reflection layer and the transmission layer iteratively, and IBCLN~\cite{IBCLN} further introduced recurrent neural networks, which is also exploited in \cite{DMGN, Location-aware}.
Prasad \etal~\cite{VDESIRR} proposed a multi-scale structure which starts from a low-resolution input and progressively grows into the desired resolution.
It is worth noting that the backbones in this paper can cover most of these architectures.

\subsection{Single-Image Reflection Removal Datasets}
\label{sec:related_datasets}
In the early studies on learning-based SIRR, due to the lack of real-world training images, synthesis methods were introduced to train the SIRR models~\cite{CEILNet, BDN}.
Subsequent works also tried to synthesize more realistic reflection-contaminated images by considering camera pose~\cite{Zhang}, generative adversarial networks~\cite{RMNet}, rendering software~\cite{physical}, and physical models~\cite{absorption}.

Meanwhile, several real-world datasets have been collected.
Zhang \etal~\cite{Zhang}, Li \etal~\cite{IBCLN}, and Wei \etal~\cite{ERRNet} captured three real-world datasets with 109, 220, and 450 image pairs, respectively.
For testing, a real-world benchmark dataset termed as \textit{SIR$^2$} has been collected~\cite{SIR2}, which is composed of three subsets, \ie, \textit{Wild}, \textit{Solid}, and \textit{Postcard}.
Interestingly, Wan \etal~\cite{CoRRN} collected 3,250 reflection layers by putting a black piece of paper behind the glass, which provides another way for synthesis.
Lei \etal~\cite{CDR} collected a perfectly aligned testing dataset by capturing raw images and obtained the transmission layer by subtracting the reflection layer from the reflection-contaminated image.

In summary, multiple datasets have been built with diverse reflection types and scenes, and the non-negligible domain gaps make it a critical problem to generalize and adapt to various domains for enhancing SIRR.


%

\subsection{Domain Generalization}
\label{sec:related_DG}
Domain generalization (DG), also known as out-of-distribution (OOD) generalization, aims at learning a model for generalizing to OOD samples, has attracted upsurging attention since it was introduced by Blanchard \etal~\cite{DG_first}.
Recent DG methods improve generalization from one or more aspects of data manipulation, representation learning, and learning strategies.
The category most relevant to this paper in DG is ensemble learning~\cite{ding2017deep, mancini2018best, d2018domain, wang2020dofe, DAEL}.
These methods generally train domain-specific branches or networks, and fuse the features~\cite{d2018domain, wang2020dofe} or predictions~\cite{mancini2018best, DAEL} via a predicted weight during inference.
Ding \etal~\cite{ding2017deep} learned a domain-invariant model by distilling domain-specific models under structured low-rank supervision.
Most DG methods focus on high-level vision tasks, in this paper, we explore the SIRR task from a domain generalization perspective, hoping to encourage the application of DG methods in low-level vision tasks.
Please refer to \cite{DG_survey_Wang, DG_survey_Zhou} for a comprehensive review of DG methods.

\subsection{Network Interpolation}
\label{sec:related_NI}
Checkpoint ensemble~\cite{checkpoint_ensemgble} is a commonly used network interpolation strategy in recent years, which interpolates the parameters from different iterations for more stable training and enhanced performance.
Chen \etal~\cite{DynamicNet} proposed to learn the residual of parameters in a pre-trained model with a different yet related learning objective, and performed model interpolation in the objective-space.
Wang \etal~\cite{wang2019deep} analyzed the interpolation between two separately trained networks, and achieved continuous imagery effect transition.
However, the interpolation weights of these methods should be manually tuned. In this paper, we leverage the power of domain generalization and predict the interpolation weights for generalizing to unseen domains.

\section{Method}
\label{sec:method}
In this section, we first provide an theoretical analysis and show the domain gaps statistically, which indicates the rationality and feasibility to solve the SIRR problem from a domain generalization perspective.
Then, given domain experts of various reflection types, a reflection type-aware weighting (RTAW) module is deployed for predicting the expert-wise weights, and we propose an in-domain expert loss for training RTAW.
An adaptive network combination (AdaNEC) method is then incorporated with RTAW for generalizing to unknown target domains.

\begin{table}[t]
  \caption{Domain classification on ERRNet training sets.}\vspace{-2mm}
  \label{tab:DG_classification}
  \centering
  \scalebox{0.9}{
  \begin{tabular}{lcccc}
    \toprule
    Dataset          & \textit{Syn$_\mathit{CEIL}$} & \textit{Real89} & \textit{Unaligned} & Total \\
    \midrule
    Training Samples & 200   & 71    & 200   & 471   \\
    Testing Samples  & 50    & 18    & 50    & 118   \\
    \midrule
    Accuracy         & 88\%  & 61\%  & 92\%  & 86\%  \\
    \bottomrule
  \end{tabular}}\vspace{-2mm}
\end{table}

\subsection{Domain Gaps between SIRR Datasets}
\label{sec:method_analysis}
In this paper, we propose to improve SIRR from a DG perspective.
Then, here comes a premise question, \ie, \textit{to what extent are these datasets different from each other}?
To answer this question, we follow Torralba \etal\cite{torralba2011unbiased} and design a simple experiment to statistically evaluate the domain gaps between different datasets.
Specifically, the training sets employed in ERRNet~\cite{ERRNet}, \ie, \textit{Syn$_\mathit{CEIL}$}~\cite{CEILNet}, \textit{Real89}~\cite{Zhang}, and \textit{Unaligned}~\cite{ERRNet}, are randomly split into training and testing sets at a ratio of around 8:2.
Then the pairs of reflection-contaminated image and the corresponding dataset index label are used to train a small-scale classification model for telling which dataset a testing image is from.
The classification model is composed of five convolution layers and one fully-connected layer.
As shown in \cref{tab:DG_classification}, the model can achieve 86\% overall accuracy.
Even with only 71 training samples, it still achieves $\sim$60\% accuracy on \textit{Real89}~\cite{Zhang} dataset.
The result clearly shows the domain gaps between SIRR datasets, making it a promising way to apply domain generalization for enhancing SIRR methods.

However, as shown in \cite{DG_survey_Zhou, DG_survey_Wang}, most DG methods focus on high-level vision tasks like image classification and semantic segmentation.
Therefore, we provide a theoretical analysis to discuss the feasibility of leveraging DG methods for SIRR.
As revealed in \cite{optical} and \cite{absorption}, the acquisition of reflection-contaminated images can be formulated by
\begin{equation}
    \setlength{\abovedisplayskip}{3mm plus 0.5ex}
    \setlength{\belowdisplayskip}{3mm plus 0.5ex}
    \mathbf{I}=\mathcal{I}(\mathbf{\Omega}\mathbf{T}^\mathit{r}+\mathbf{\Phi}\mathbf{R}^\mathit{r}),
    \label{eqn:isp}
\end{equation}
where $\mathbf{T}^\mathit{r}$ and $\mathbf{R}^\mathit{r}$ are the raw signals of transmission and reflection layers, $\mathbf{\Omega}$ and $\mathbf{\Phi}$ represent the refractive and reflective amplitude coefficient maps relevant to the glass characteristics and photographing parameters, and $\mathcal{I}$ denotes the camera image signal processing (ISP) pipeline to obtain sRGB images.
Most synthesis methods~\cite{CEILNet, BDN, Zhang, physical, absorption} also approximate this formulation.
In other words, considering the parameters in \cref{eqn:isp}, a manifold is spanned by the images with diverse reflection types, which settles the foundation of our DG-based solution.

\begin{figure}[t]
    \centering
    \begin{overpic}[width=.95\linewidth]{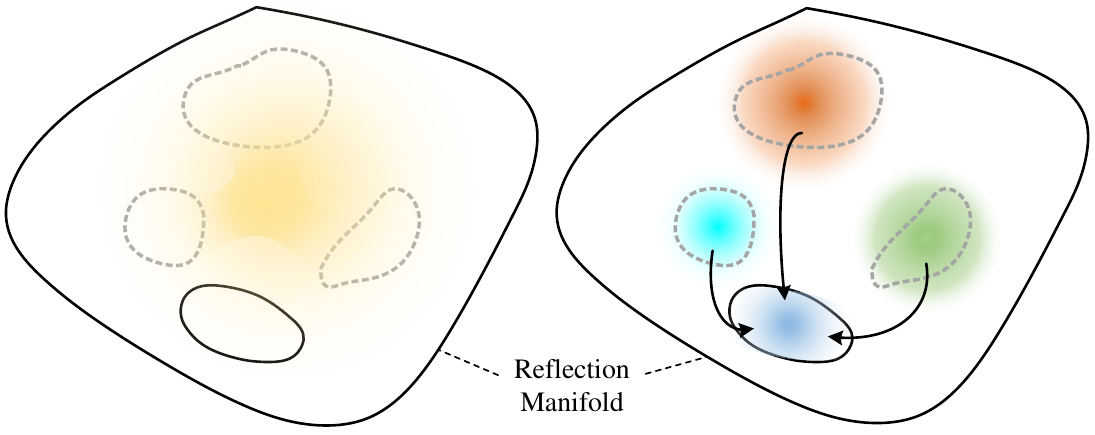}
        \put(22, 29){$\mathcal{S}_1$}
        \put(13.2, 17.5){$\mathcal{S}_2$}
        \put(32.2, 16){$\mathcal{S}_3$}
        \put(21, 8){$\mathcal{T}$}

        \put(72, 29){$\mathcal{S}_1$}
        \put(63.2, 17.5){$\mathcal{S}_2$}
        \put(82.2, 16){$\mathcal{S}_3$}
        \put(71, 8){$\mathcal{T}$}

        \put(71.5, 19){$\mathit{w}_1$}
        \put(60, 11){$\mathit{w}_2$}
        \put(82, 7){$\mathit{w}_3$}

    \end{overpic}

    \begin{tabular}{p{0.45\linewidth}<{\centering}p{0.45\linewidth}<{\centering}}
        (a) Joint-training & (b) Adaptive Combination
    \end{tabular}
    \caption{Comparison on SIRR working schemes. (a)~The joint-training scheme of previous methods leads to a limited and fixed supported domain. (b)~The proposed adaptive network combination method can generalize to the target domain adaptively during inference.
    The circle means the reflection manifold or a subspace in it, and the color area (strength) represents the supporting range (expertise level). $\mathcal{S}_\mathit{i}$ and $\mathcal{T}$ denote the $\mathit{i}$-th training set (source domain) and a testing set (target domain), respectively.}
    \label{fig:training_scheme}
\end{figure}
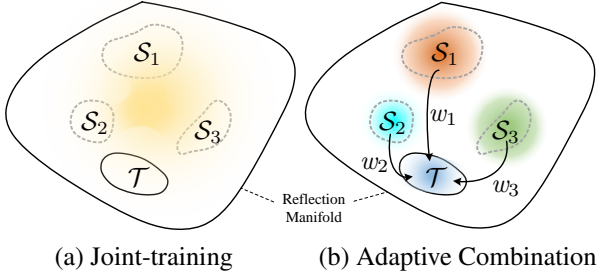

\subsection{Overall Solution}
\label{sec:method_overall}
Since most previous SIRR methods are trained with a mixture of various datasets, they have to compromise between different domains, which is often not within the scope of target testing sets (see \cref{fig:training_scheme}(a)).
Instead, suppose that there are $\mathit{N}$ training datasets, where the $\mathit{i}$-th dataset is denoted by $\mathcal{S}_\mathit{i}$, we train an SIRR model for each of the dataset.
In this case, we can obtain $\mathit{N}$ domain experts $\{\mathit{G}_\mathit{i}\}_{\mathit{i}=1}^\mathit{N}$ (see the \textit{Red}, \textit{Cyan} and \textit{Green} regions in \cref{fig:training_scheme}(b)), where the expert $\mathit{G}_\mathit{i}$ is specialized on the relevant reflection types in $\mathcal{S}_\mathit{i}$.
%
Then, for a reflection-contaminated image $\mathbf{I}$, we can combine these domain experts for better reflection removal according to certain weights (\eg, $\mathit{w}_1\!\sim\!\mathit{w}_3$ in \cref{fig:training_scheme}(b)).
%
%
A reflection type-aware weighting (RTAW) module is introduced to evaluate the expertise level and predict the expert-wise weights of each model $\mathit{G}_\mathit{i}$ on $\mathbf{I}$, where an adaptive network combination (AdaNEC) method is incorporated to better leverage the predicted RTAW weights.
In the following, we explain the design of RTAW and AdaNEC in detail.


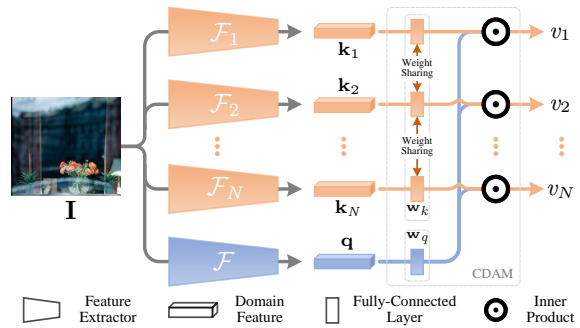
\begin{figure}[t]
    \centering
  \begin{overpic}[width=.95\linewidth]{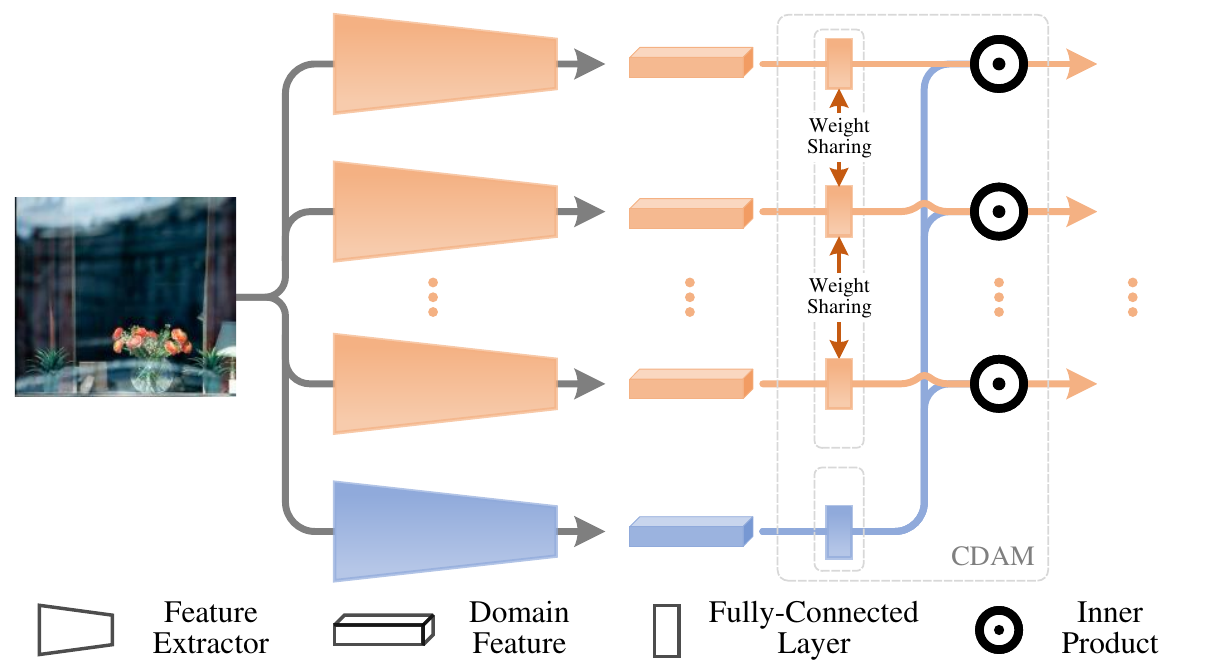}
    \put(10.25, 18.5){$\mathbf{I}$}
    \put(34.5,48.5){\normalsize {\color{white}$\mathcal{F}_1$}}
    \put(34.5,36.5){\normalsize {\color{white}$\mathcal{F}_2$}}
    \put(34,22.5){\normalsize {\color{white}$\mathcal{F}_\mathit{N}$}}
    \put(35,10){\normalsize {\color{white}$\mathcal{F}$}}



    \put(67.5,19.9){\tiny $\mathbf{w}_\mathit{k}$}
    \put(67.5,15.25){\tiny $\mathbf{w}_\mathit{q}$}

    \put(56, 46){\scriptsize $\mathbf{k}_1$}
    \put(56, 40){\scriptsize $\mathbf{k}_2$}
    \put(55.5, 19.5){\scriptsize $\mathbf{k}_\mathit{N}$}
    \put(56.5, 14){\scriptsize $\mathbf{q}$}

    \put(91.5, 49){\small $\mathit{v}_1$}
    \put(91.5, 36.9){\small $\mathit{v}_2$}
    \put(91, 22.5){\small $\mathit{v}_N$}
  \end{overpic}
  \caption{Structure of reflection type-aware weighting (RTAW) module. $\mathbf{I}$ denotes a reflection-contaminated image (the shown image is collected by Fan \etal \cite{CEILNet}), $\mathbf{k}_\mathit{i}$ and $\mathbf{q}$ are vectors extracted by $\mathcal{F}_\mathit{i}$ and $\mathcal{F}$, $\mathit{v}_i$ means expertise level of the $\mathit{i}$-th expert $\mathit{G}_\mathit{i}$ on image $\mathbf{I}$, which can be used to calculate the expert-wise RTAW weights (\eg, $\mathit{w}_\mathit{i}$ as shown in \cref{fig:training_scheme}(b)).}
  \label{fig:RTAW}
\end{figure}

\subsection{Reflection Type-aware Weighting Module}
\label{sec:method_RTAW}
The reflection type-aware weighting (RTAW) module is comprised of two parts, \ie, the feature extraction networks and a cross-domain attention module.
As shown in \cref{fig:RTAW}, a feature extractor $\mathcal{F}$ is deployed to extract the universal feature vector $\mathbf{q}$ from the reflection-contaminated input $\mathbf{I}$, \ie, $\mathbf{q}=\mathcal{F}(\mathbf{I})$.
Then, for each domain expert $\mathit{G}_\mathit{i}$, we obtain a domain-specific feature vector $\mathbf{k}_\mathit{i}$ via the corresponding extractor $\mathcal{F}_\mathit{i}$, \ie, $\mathbf{k}_\mathit{i}=\mathcal{F}_\mathit{i}(\mathbf{I})$.
Inspired by the self-attention layer in Transformer~\cite{transformer}, we present a cross-domain attention module (CDAM) to evaluate the expertise level of each domain expert by matching their feature vector $\mathbf{k}_\mathit{i}$ with the common feature $\mathbf{q}$, \ie,
\begin{equation}
    \setlength{\abovedisplayskip}{2mm plus 0.5ex}
    \setlength{\belowdisplayskip}{2mm plus 0.5ex}
    \mathit{v}_\mathit{i}=\left(\mathbf{W}_\mathit{k}^\top\mathbf{k}_\mathit{i}\right)\odot\left(\mathbf{W}_\mathit{q}^\top\mathbf{q}\right),
\end{equation}
where $\mathbf{W}_\mathit{k}$ and $\mathbf{W}_\mathit{q}$ are parameters of the fully-connected layers for $\mathbf{k}$ and $\mathbf{q}$ respectively, and $\mathbf{W}_\mathit{k}$ is shared by all of $\mathbf{k}_1$ to $\mathbf{k}_\mathit{N}$, $\odot$ denotes inner-product.
Then for a testing image, all domain experts are exploited, while the RTAW weights can be calculated by using the expertise level, \ie,
\begin{equation}
    \setlength{\abovedisplayskip}{2mm plus 0.5ex}
    \setlength{\belowdisplayskip}{2mm plus 0.5ex}
    \mathbf{w}=\sigma(\mathbf{v}),
\end{equation}
where $\sigma$ is the softmax operation and $\mathbf{v}=\{\mathit{v}_\mathit{i}\}_{\mathit{i}=1}^\mathit{N}$.

\subsection{Learning Objective}
\label{sec:method_loss}
Apart from the reflection removal losses, an in-domain expert (IDE) loss is introduced for training the proposed RTAW module, and we introduce the two parts of the learning objective in the following.
Note that parameters of the domain experts are fixed when training the RTAW module.

\vspace{0.5em}
\noindent\textbf{Reflection Removal Losses.}
When training RTAW, there is no extra dataset to serve as the target domain, thus we follow the leave-one-domain-out strategy~\cite{LODO}.
Specifically, given a training sample $\mathbf{I}_\mathit{tr}$ from the $\mathit{i}$-th training set $\mathcal{S}_\mathit{i}$, we denote by $\mathbf{T}$ and $\mathbf{R}$ the ground-truth transmission and reflection layers, respectively. The $\mathit{i}$-th source domain will serve as a pseudo target domain during this iteration.
Only domain experts except for $\mathit{G}_\mathit{i}$ will be employed, \ie,
\begin{align}
    \setlength{\abovedisplayskip}{1mm plus 0.5ex}
    \setlength{\belowdisplayskip}{1mm plus 0.5ex}
    \mathbf{w}_\mathit{i}^{_\complement}&=\sigma(\mathbf{v}_\mathit{i}^{_\complement}),\\
    \mathbf{\hat{T}}_\mathit{i}^{_\complement}&={\sum}_{\mathit{j}\neq\mathit{i}}\mathit{w}_\mathit{j}\cdot\mathit{G}_\mathit{j}(\mathbf{I}_\mathit{tr}),
\end{align}
where {\footnotesize$\complement$} denotes the complement set and $\mathbf{v}_\mathit{i}^{_\complement}=\{\mathit{v}_\mathit{j}\}_{\mathit{j}\neq\mathit{i}}$, $\mathbf{w}_\mathit{i}^{_\complement}=\{\mathit{w}_\mathit{j}\}_{\mathit{j}\neq\mathit{i}}$.
%
%
%
Note that the reflection layer $\mathbf{\hat{R}}_\mathit{i}^{_\complement}$ is generated in a similar way if the backbone model requires it for loss calculation.
In the following, we briefly summarize the reflection removal losses.

\textbf{\textit{Fidelity Loss.}}
Generally $\ell_1$ or $\ell_2$ is employed as reconstruction loss on the transmission layer or reflection layer,
\begin{equation}
    \setlength{\abovedisplayskip}{2mm plus 0.5ex}
    \setlength{\belowdisplayskip}{2mm plus 0.5ex}
    \mathcal{L}_\mathit{fid}=\left\|\mathbf{\hat{Y}}^{_\complement}_\mathit{i}-\mathbf{Y}\right\|_\mathit{p}, \mathbf{Y}\in\{\mathbf{T}, \mathbf{R}, \mathbf{\nabla\mathbf{T}}\},
\end{equation}
where $\nabla$ means gradient calculation operation, $\mathit{p}\in\{1,2\}$, and $\mathbf{Y}$ denotes the corresponding ground-truth of $\mathbf{\hat{Y}}^{_\complement}_\mathit{i}$.

\textbf{\textit{Reconstruction Loss.}}
For synthetic pairs, the synthesis methods can be leveraged for better training, and the reconstruction loss is defined as
\begin{equation}
    \setlength{\abovedisplayskip}{2mm plus 0.5ex}
    \setlength{\belowdisplayskip}{2mm plus 0.5ex}
    \mathcal{L}_\mathit{rec}=\left\|Syn(\mathbf{\hat{T}}_\mathit{i}^{_\complement}, \mathbf{\hat{R}}_\mathit{i}^{_\complement}) - \mathbf{I}_\mathit{tr}\right\|_\mathit{p},
\end{equation}
where $Syn$ denotes the differentiable synthesis methods.

\textbf{\textit{VGG-based Loss.}}
The VGG~\cite{VGG} architecture has been widely used for loss function design, \ie,
\begin{equation}
    \setlength{\abovedisplayskip}{2mm plus 0.5ex}
    \setlength{\belowdisplayskip}{2mm plus 0.5ex}
    \mathcal{L}_\mathit{VGG}={\sum}_\mathit{l}\lambda_\mathit{l}\left\|(\phi_\mathit{l}(\mathbf{\hat{T}}_\mathit{i}^{_\complement}) - \phi_\mathit{l}(\mathbf{T})\right\|_\mathit{p},
\end{equation}
where $\phi_\mathit{l}$ denotes the $\mathit{l}$-th layer of the VGG model, $\lambda_\mathit{l}$ is the corresponding weight.
The VGG-based loss is generally used following perceptual loss~\cite{Perceptual} ($\mathcal{L}_\mathit{percept}$).
In \cite{ERRNet}, Wei \etal proposed to use the ``conv5\_2'' feature for alignment-invariant loss ($\mathcal{L}_\mathit{inv}$).

\textbf{\textit{Adversarial Loss.}}
Adversarial loss~\cite{GAN} is an effective approach to enhancing the output image quality.
Given a discriminator $\mathit{D}$, the adversarial loss can be defined by
\begin{equation}
    \setlength{\abovedisplayskip}{2mm plus 0.5ex}
    \setlength{\belowdisplayskip}{2mm plus 0.5ex}
    \mathcal{L}_\mathit{adv}=-\log\mathit{D}(\mathbf{\hat{T}}_\mathit{i}^{_\complement}).
\end{equation}

To sum up, the reflection removal loss $\mathcal{L}_\mathit{RR}$ can be defined by combining these loss functions or their variants,
\begin{equation}
    \setlength{\abovedisplayskip}{2mm plus 0.5ex}
    \setlength{\belowdisplayskip}{2mm plus 0.5ex}
    \begin{aligned}
        \mathcal{L}_\mathit{RR}=&\lambda_\mathit{fid}\mathcal{L}_\mathit{fid}+\lambda_\mathit{rec}\mathcal{L}_\mathit{rec}+\lambda_\mathit{percept}\mathcal{L}_\mathit{percept}\\+&\lambda_\mathit{inv}\mathcal{L}_\mathit{inv}+\lambda_\mathit{adv}\mathcal{L}_\mathit{adv},
    \end{aligned}
\end{equation}
where $\lambda_\mathit{fid}$, $\lambda_\mathit{rec}$, $\lambda_\mathit{percept}$, $\lambda_\mathit{inv}$, and $\lambda_\mathit{adv}$ are hyper parameters following the original settings in the backbone methods. Note that they can be zero if the corresponding loss function is not applied in a backbone method.


\vspace{0.5em}
\noindent\textbf{In-domain Expert Loss.}
Suppose that when the $\mathit{i}$-th source domain $\mathcal{S}_\mathit{i}$ serves as the pseudo target domain, \textit{what if the domain expert $\mathit{G}_\mathit{i}$ is also deployed?}
Since $\mathit{G}_\mathit{i}$ is trained on $\mathcal{S}_\mathit{i}$, it can be regarded as an in-domain expert.
Therefore, we can require that the final results mostly rely on it.
To this end, we propose an in-domain expert (IDE) loss, which is defined based on the cross-entropy (CE) loss~\cite{crossentropy,mannor2005cross} by
\begin{equation}
    \setlength{\abovedisplayskip}{2mm plus 0.5ex}
    \setlength{\belowdisplayskip}{2mm plus 0.5ex}
    \mathcal{L}_\mathit{IDE}=CE(\mathbf{w}, \mathit{i})=-\log\mathit{w}_\mathit{i}.
\end{equation}
We empirically find that the IDE loss not only promotes the reflection removal performance but also makes the training more stable.
Further discussions are given in \cref{sec:ablation_study}.

The overall learning objective for training RTAW is
\begin{equation}
    \setlength{\abovedisplayskip}{2mm plus 0.5ex}
    \setlength{\belowdisplayskip}{2mm plus 0.5ex}
    \mathcal{L}=\mathcal{L}_\mathit{RR}+\lambda\mathcal{L}_\mathit{IDE},
\end{equation}
where $\lambda$ is set to 0.1 for all three backbone methods.
Please refer to the supplementary material for more details about the loss function configuration in this paper.

\begin{table}[t]
    \caption{Dataset configuration of the backbone methods.}\vspace{-2mm}
    \label{tab:exp_datasets}
    \centering
    \scalebox{0.75}{
        \begin{tabular}{lcc}
            \toprule
            \multicolumn{1}{c}{Backbone} & Training Datasets & Testing Datasets \\
            \midrule
            ERRNet~\cite{ERRNet} & \textit{Syn$_\mathit{CEIL}$}, \textit{Real89}, \textit{Unaligned} & \textit{Real20}, \textit{SIR$^2$} \\
            IBCLN~\cite{IBCLN} & \textit{Syn}$_\mathit{Zhang}$, \textit{Real89}, \textit{Nature200} & \textit{Real20}, \textit{SIR$^2$}, \textit{Nature20} \\
            RAGNet~\cite{RAGNet} & \textit{Syn$_\mathit{CEIL}$}, \textit{Syn$_\mathit{Zhang}$}, \textit{Real89} & \textit{Real20}, \textit{SIR$^2$} \\
            \bottomrule
    \end{tabular}}
\end{table}

\subsection{Adaptive Network Combination}
\label{sec:method_AdaNEC}
As shown in \cref{fig:training_scheme}(b), the RTAW weights can be employed for adaptively generalizing to target domains via adaptive network combination (AdaNEC). In this paper, considering both adaptation level and efficiency, we provide two representative AdaNEC methods termed output fusion (OF) and network interpolation (NI), respectively.

\vspace{0.5em}
\noindent\textbf{Output Fusion (OF).}
The OF manner is similar to the training procedure of RTAW, except that all domain experts are employed during inference to generate the output, \ie,
\begin{equation}
    \setlength{\abovedisplayskip}{3mm plus 0.5ex}
    \setlength{\belowdisplayskip}{3mm plus 0.5ex}
    \mathbf{\hat{T}}_\mathit{OF}={\sum}_\mathit{i=1}^\mathit{N}\mathit{w}_\mathit{i}\cdot\mathit{G}_\mathit{i}(\mathbf{I}),
\end{equation}
where the results generated by the domain experts are fused as final output.
However, such a method requires multiple model inference, which brings heavy extra computation burden and time consumption.

\vspace{0.5em}
\noindent\textbf{Network Interpolation (NI).}
To alleviate this problem, the NI manner interpolates the model parameters in advance, and the inference procedure is processed with the fused model parameters, which can be formulated by
\begin{equation}
    \setlength{\abovedisplayskip}{2mm plus 0.5ex}
    \setlength{\belowdisplayskip}{2mm plus 0.5ex}
    \mathbf{\hat{T}}_\mathit{NI}=\left({\sum}_\mathit{i=1}^\mathit{N}\mathit{w}_\mathit{i}\cdot\mathit{G}_\mathit{i}\right)(\mathbf{I}),
\end{equation}
where only a few extra computations (\ie, the lightweight RTAW module and parameter interpolation) are introduced.

Besides, as shown in \cref{sec:method_analysis}, generally the reflection-contaminated images in a dataset form a domain.
In this case, we can use the average RTAW weights of a dataset for more stable prediction and apply domain-level AdaNEC.

\begin{table*}[t]
    \caption{Quantitative comparison against the state-of-the-art backbones. For calculating the PSNR and SSIM indices, we follow the testing protocol of the backbone methods. Note that the results of ERRNet~\cite{ERRNet} and RAGNet~\cite{RAGNet} on \textit{Nature20} are given for reference (highlighted with gray shadow), and the average indices are calculated on the other four datasets (\textit{Real20}, \textit{Wild}, \textit{Postcard}, and \textit{Solid}) according to the official setting. The best results are marked with \textbf{bold}, and the image amount of each dataset is provided in the parentheses.}\vspace{-1mm}
    \label{tab:exp_main}
    \centering
    \scalebox{0.88}{
    \begin{tabular}{lcccccccccccc}
        \toprule
        \multicolumn{1}{c}{}                          & \multicolumn{2}{c}{\textit{Real20} (20)}      & \multicolumn{2}{c}{\textit{Wild} (55)}        & \multicolumn{2}{c}{\textit{Postcard} (199)}    & \multicolumn{2}{c}{\textit{Solid} (200)}       & \multicolumn{2}{c}{\textit{Nature20} (20)}                                                    & \multicolumn{2}{c}{Average (474/494)}     \\
        \cmidrule(lr){2-3}  \cmidrule(lr){4-5} \cmidrule(lr){6-7}  \cmidrule(lr){8-9}  \cmidrule(lr){10-11}  \cmidrule(lr){12-13}
        \multicolumn{1}{c}{\multirow{-2}{*}{Methods}} & PSNR$\uparrow$ & SSIM$\uparrow$ & PSNR$\uparrow$ & SSIM$\uparrow$ & PSNR$\uparrow$ & SSIM$\uparrow$ & PSNR$\uparrow$ & SSIM$\uparrow$ & PSNR$\uparrow$                         & SSIM$\uparrow$                         & PSNR$\uparrow$ & SSIM$\uparrow$ \\ \midrule \midrule
        ERRNet~\cite{ERRNet}                          & 22.08          & 0.781          & 25.13          & 0.889          & 22.76          & 0.864          & 24.62          & 0.898          & \cellcolor[HTML]{EFEFEF}20.86          & \cellcolor[HTML]{EFEFEF}0.757          & 23.79          & 0.877          \\
        ERRNet$_\mathit{OF}$                          & 22.80          & 0.790          & 25.26          & 0.890          & 23.08          & 0.874          & \textbf{25.26} & \textbf{0.904} & \cellcolor[HTML]{EFEFEF}20.99          & \cellcolor[HTML]{EFEFEF}0.768          & 24.24          & 0.885          \\
        ERRNet$_\mathit{NI}$                          & \textbf{22.81} & \textbf{0.791} & \textbf{25.70} & \textbf{0.895} & \textbf{23.56} & \textbf{0.884} & 25.13          & 0.902          & \cellcolor[HTML]{EFEFEF}\textbf{21.20} & \cellcolor[HTML]{EFEFEF}\textbf{0.771} & \textbf{24.44} & \textbf{0.889} \\ \midrule
        IBCLN~\cite{IBCLN}                            & 21.86          & 0.762          & 24.71          & 0.886          & 23.39          & 0.875          & 24.87          & 0.893          & 24.03                                  & 0.787                                  & 24.10          & 0.877          \\
        IBCLN$_\mathit{OF}$                           & \textbf{22.52} & \textbf{0.789} & \textbf{25.77} & \textbf{0.897} & \textbf{24.27} & \textbf{0.889} & \textbf{25.24} & \textbf{0.900} & \textbf{24.74}                         & \textbf{0.820}                         & \textbf{24.78} & \textbf{0.888} \\
        IBCLN$_\mathit{NI}$                           & 22.04          & 0.782          & 25.35          & 0.894          & 23.34          & 0.887          & 24.85          & 0.897          & 24.59                                  & 0.818                                  & 24.17          & 0.885          \\ \midrule
        RAGNet~\cite{RAGNet}                          & 22.95          & 0.793          & 25.52          & 0.880          & 23.67          & 0.879          & 26.15          & 0.903          & \cellcolor[HTML]{EFEFEF}21.21          & \cellcolor[HTML]{EFEFEF}0.765          & 24.90          & 0.886          \\
        RAGNet$_\mathit{OF}$                          & \textbf{23.34} & \textbf{0.807} & 25.85          & 0.896          & \textbf{25.20} & \textbf{0.903} & \textbf{26.17} & \textbf{0.908} & \cellcolor[HTML]{EFEFEF}\textbf{21.48} & \cellcolor[HTML]{EFEFEF}0.776          & \textbf{25.60} & \textbf{0.900} \\
        RAGNet$_\mathit{NI}$                          & 23.18          & 0.802          & \textbf{26.25} & \textbf{0.899} & 24.90          & 0.906          & 25.66          & 0.903          & \cellcolor[HTML]{EFEFEF}21.44          & \cellcolor[HTML]{EFEFEF}\textbf{0.777} & 25.31          & \textbf{0.900} \\ \bottomrule
    \end{tabular}}
\end{table*}

\section{Experiments}
\label{sec:experiments}

\subsection{Implementation Details}
\label{sec:exp_implementation}

\noindent\textbf{Datasets.}
Since the core idea of this work is to make better use of the training sets and enhance SIRR from a domain generalization perspective, we follow the training configuration of the backbone methods and have no requirements on extra training datasets.
In the following, the relevant training and testing datasets are briefly introduced.

\textit{\textbf{Syn$_\mathit{\textbf{CEIL}}$}} - The synthetic dataset proposed in CEILNet~\cite{CEILNet}, where 7,643 image pairs are selected from the PASCAL VOC\footnote{\url{http://host.robots.ox.ac.uk/pascal/VOC/}} dataset~\cite{VOC}, and each pair serves as transmission and reflection layer, respectively.
Blurriness and luminance of the reflection layer are considered for synthesis.

\textit{\textbf{Syn$_\mathit{\textbf{Zhang}}$}} - The synthetic dataset proposed by Zhang \etal~\cite{Zhang}, where 13,700 indoor-outdoor image pairs are collected from Flickr\footnote{\url{https://www.flickr.com/}}. The synthesis method is adapted from \textit{Syn$_\mathit{CEIL}$} and considers the angle for capturing images.

\textit{\textbf{Real89 \& Real20, Nature200 \& Nature20}} - Two well aligned real-world datasets captured by Zhang \etal~\cite{Zhang} and Li \etal~\cite{IBCLN}, where \textit{Real89} and \textit{Nature200} are used for training while \textit{Real20} and \textit{Nature20} are used for testing.

\textit{\textbf{Unaligned}} - The training set collected by Wei \etal~\cite{ERRNet}, where the reflection-contaminated images and transmission layers are mildly aligned. Note that there are 250 pairs captured by DSLR camera and 200 by mobile phone, we use the 250 DSLR pairs following the official code of \cite{ERRNet}.

\textit{\textbf{SIR$^2$}} - A benchmark dataset\footnote{\url{https://rose1.ntu.edu.sg/dataset/sir2Benchmark}} collected by Wan \etal~\cite{SIR2}, which is composed of three subsets, \ie, \textit{Wild} (55), \textit{Solid} (200), and \textit{Postcard} (199), where the \textit{Wild} subset is captured in the wild and the other two subsets are collected in a controlled laboratory environment.

The training/testing dataset configuration of the backbone methods is summarized in \cref{tab:exp_datasets}.

\vspace{0.5em}
\noindent\textbf{Backbones and Implementation Details.}
In this paper, we use three state-of-the-art methods as backbone models, \ie, ERRNet\footnote{\url{https://github.com/Vandermode/ERRNet}}~\cite{ERRNet}, IBCLN\footnote{\url{https://github.com/JHL-HUST/IBCLN}}~\cite{IBCLN}, and RAGNet\footnote{\url{https://github.com/liyucs/RAGNet}}~\cite{RAGNet}.
As introduced in \cref{sec:related_SIRR}, they are the best performed SIRR methods with source code and pre-trained model available, and most of recent structure design categories are covered by these methods.
All the three methods are implemented in the PyTorch~\cite{PyTorch} framework.
The experiments are conducted on a PC with an Nvidia RTX 3090 GPU.
The source code and pre-trained models will be publicly available.

\vspace{0.5em}
\noindent\textbf{Testing Protocols.}
As shown in \cref{tab:exp_datasets}, the backbone methods are trained and evaluated on various datasets.
For a fair comparison, apart from the proposed IDE Loss $\mathcal{L}_\mathit{IDE}$, we train our method with only the training sets and loss functions it used given a backbone model (please refer to the supplementary material for more details), and evaluate the model performance following its testing protocol.
Since only the model trained with aligned image pairs is given, we fine-tuned the ERRNet model following official instructions, and achieved an average PSNR index of 23.79 dB, which is higher than the performance (23.59 dB) reported by Wei~\etal~\cite{ERRNet}.
We also note that since the image size of \textit{Real20} is too large for inference, following the backbone methods, the shorter side of \textit{Real20} images is resized to 512, 400, and 300 pixels for evaluating ERRNet~\cite{ERRNet}, IBCLN~\cite{IBCLN}, and RAGNet~\cite{RAGNet}, respectively.

\begin{figure}[t]
  \small
  \centering
  \scalebox{.9}{
    \begin{tabular}{ccccc}
      \includegraphics[width=0.124\textwidth]{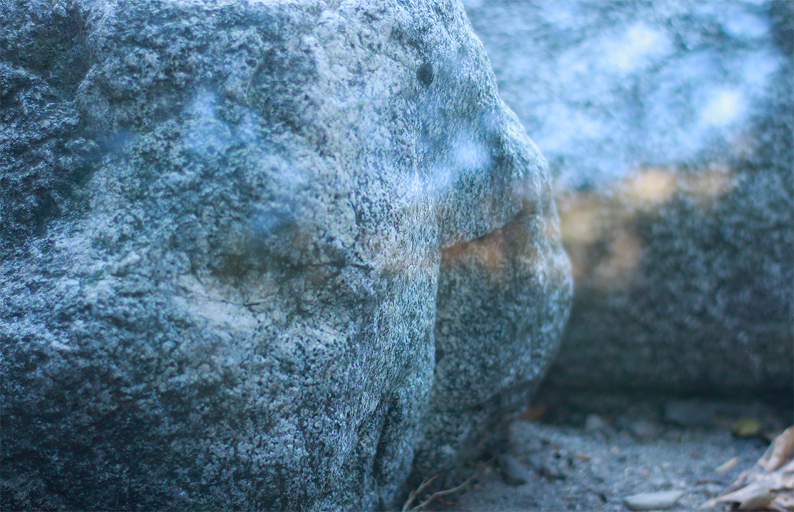}&\hspace{-4.2mm}
      \includegraphics[width=0.124\textwidth]{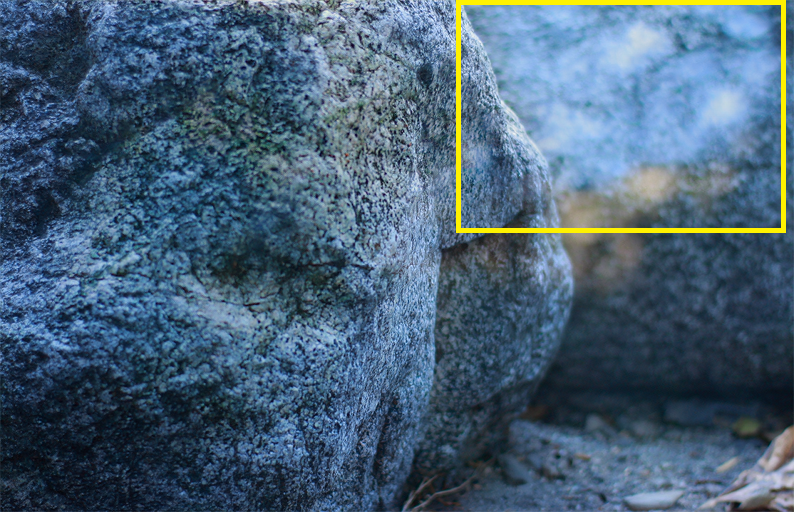}&\hspace{-4.2mm}
      \includegraphics[width=0.124\textwidth]{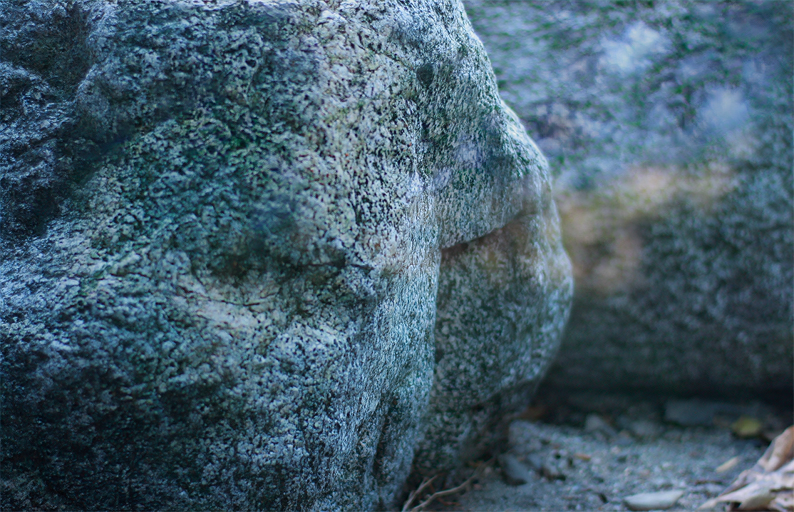}&\hspace{-4.2mm}
      \includegraphics[width=0.124\textwidth]{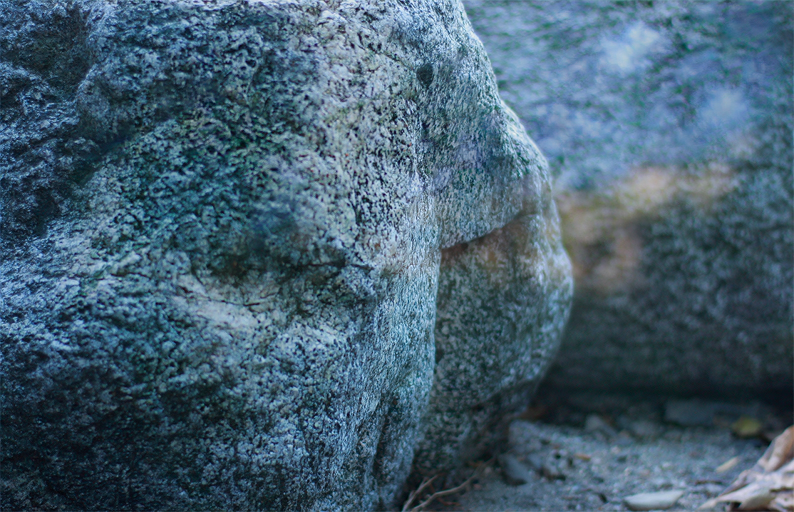}&\hspace{-4.2mm}
      \\
      \includegraphics[width=0.124\textwidth]{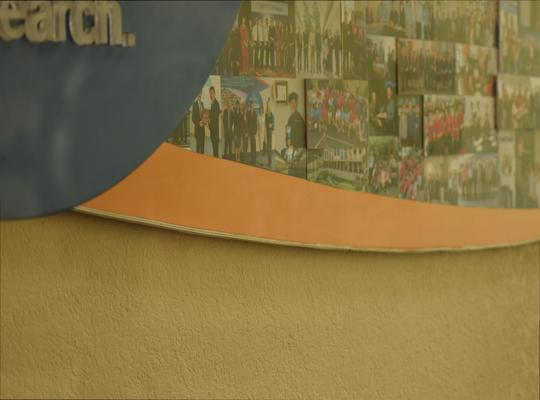}&\hspace{-4.2mm}
      \includegraphics[width=0.124\textwidth]{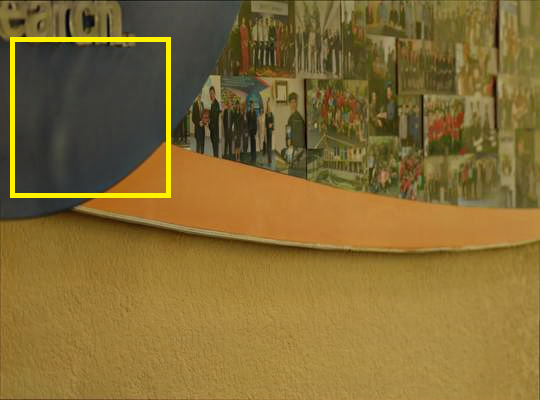}&\hspace{-4.2mm}
      \includegraphics[width=0.124\textwidth]{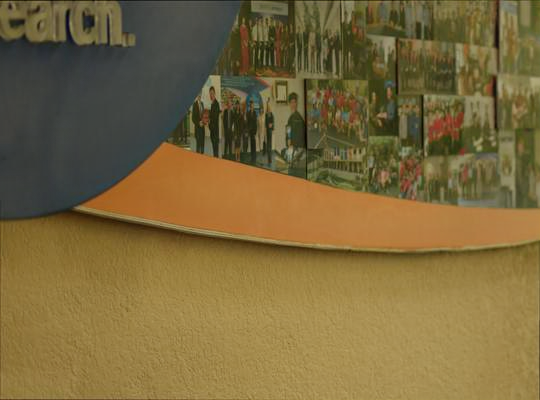}&\hspace{-4.2mm}
      \includegraphics[width=0.124\textwidth]{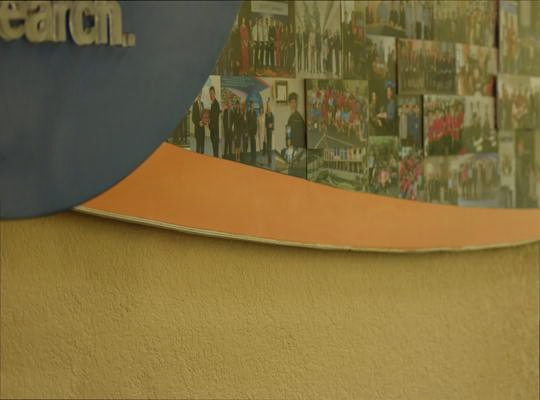}&\hspace{-4.2mm}
      \\
      Input&\hspace{-4.2mm}
      ERRNet~\cite{ERRNet}&\hspace{-4.2mm}
      ERRNet$_\mathit{OF}$&\hspace{-4.2mm}
      ERRNet$_\mathit{NI}$& \hspace{-4.2mm}
      \\
    \end{tabular}}
  \vspace{-3mm}
  \captionsetup{font={small}}
  \caption{Visual comparison of ERRNet based AdaNEC methods.}
  \label{fig:vis_errnet}
  \vspace{-2mm}
\end{figure}

\begin{figure}[t]
  \small
  \centering
    \scalebox{.9}{
      \begin{tabular}{ccccc}
        \includegraphics[width=0.124\textwidth]{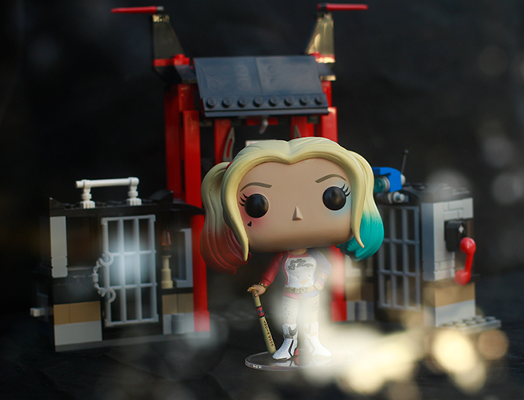}&\hspace{-4.2mm}
        \includegraphics[width=0.124\textwidth]{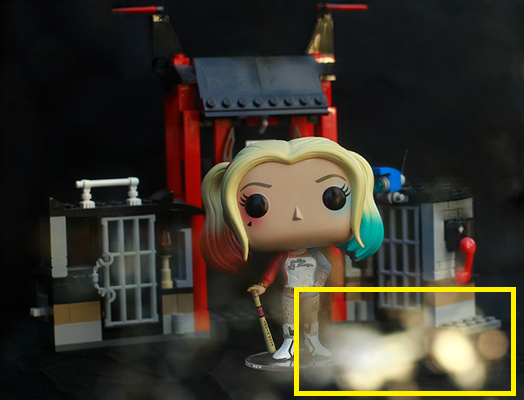}&\hspace{-4.2mm}
        \includegraphics[width=0.124\textwidth]{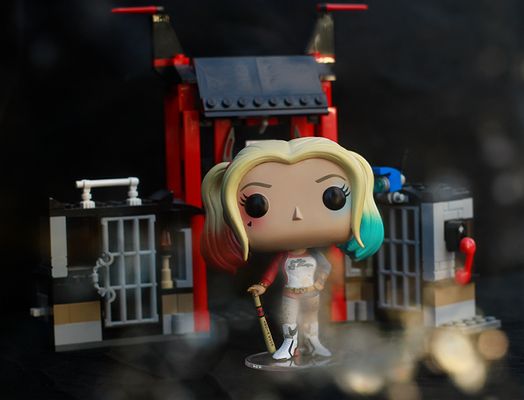}&\hspace{-4.2mm}
        \includegraphics[width=0.124\textwidth]{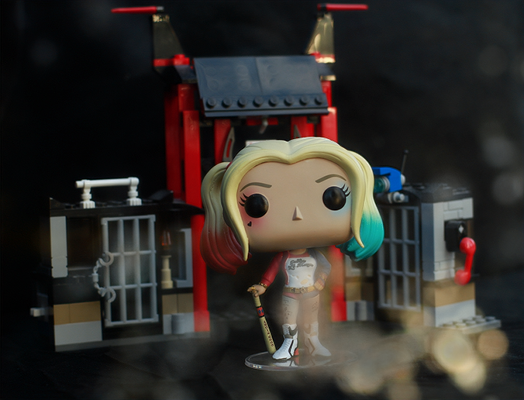}&\hspace{-4.2mm}
        \\
        \includegraphics[width=0.124\textwidth]{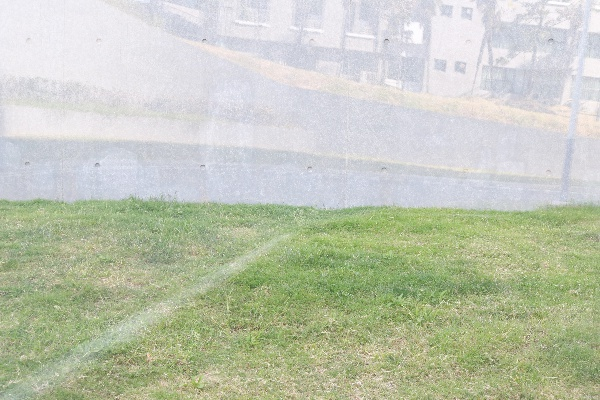}&\hspace{-4.2mm}
        \includegraphics[width=0.124\textwidth]{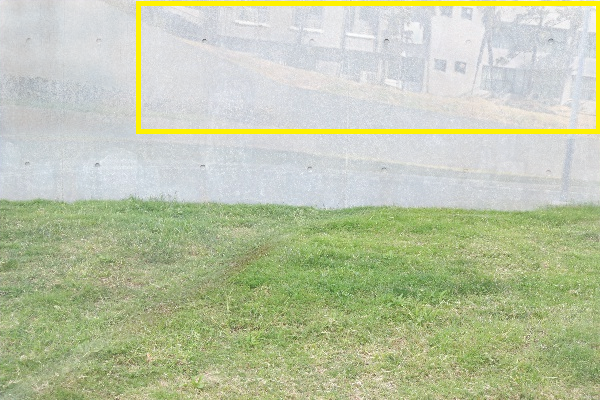}&\hspace{-4.2mm}
        \includegraphics[width=0.124\textwidth]{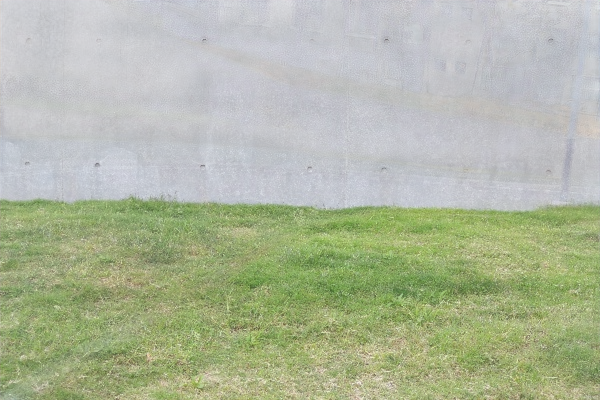}&\hspace{-4.2mm}
        \includegraphics[width=0.124\textwidth]{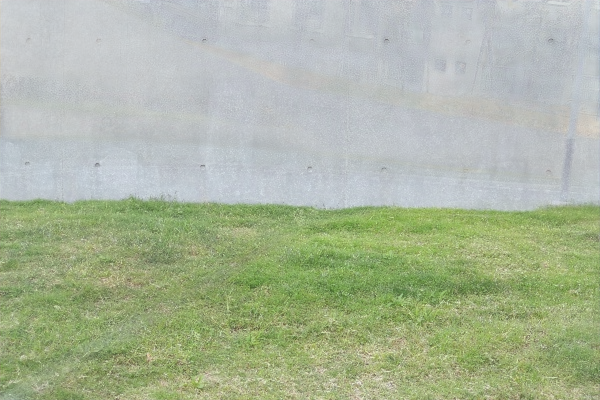}&\hspace{-4.2mm}
        \\
        Input&\hspace{-4.2mm}
        IBCLN~\cite{IBCLN}&\hspace{-4.2mm}
        IBCLN$_\mathit{OF}$&\hspace{-4.2mm}
        IBCLN$_\mathit{NI}$& \hspace{-4.2mm}
        \\
    \end{tabular}}
  \vspace{-3mm}
  \captionsetup{font={small}}
  \caption{Visual comparison of IBCLN based AdaNEC methods.}
  \label{fig:vis_ibcln}
  \vspace{-2mm}
\end{figure}

\begin{figure}[t]
  \small
  \centering
    \scalebox{.9}{
      \begin{tabular}{ccccc}
        \includegraphics[width=0.124\textwidth]{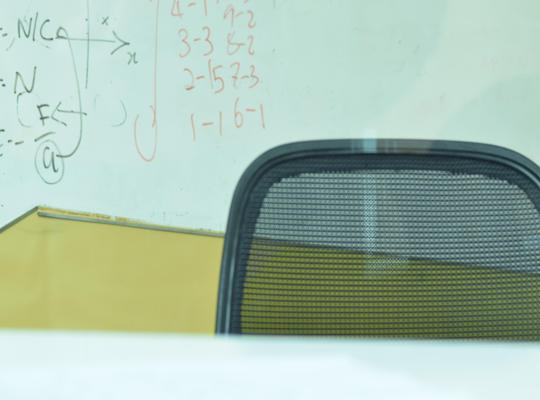}&\hspace{-4.2mm}
        \includegraphics[width=0.124\textwidth]{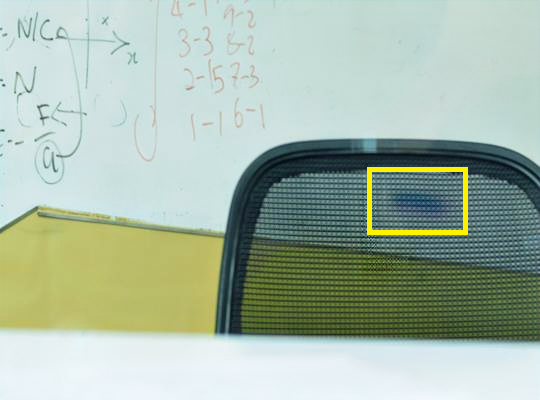}&\hspace{-4.2mm}
        \includegraphics[width=0.124\textwidth]{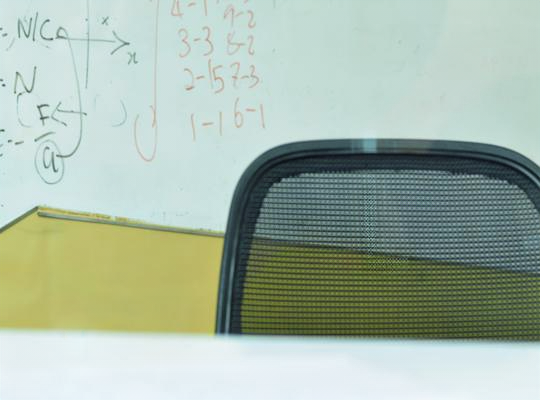}&\hspace{-4.2mm}
        \includegraphics[width=0.124\textwidth]{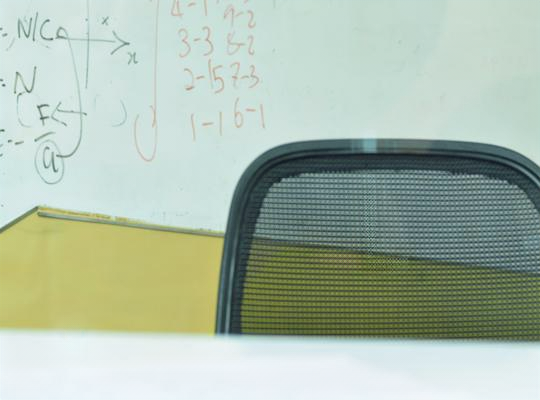}&\hspace{-4.2mm}
        \\
        \includegraphics[width=0.124\textwidth]{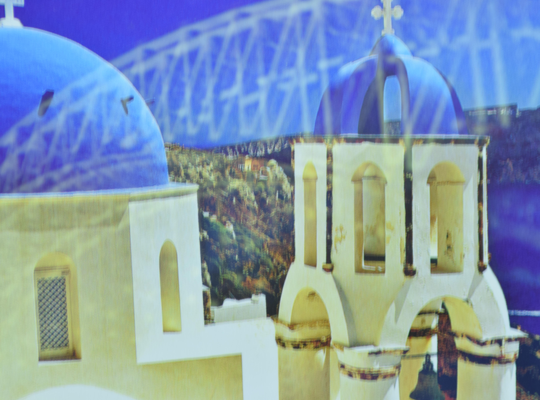}&\hspace{-4.2mm}
        \includegraphics[width=0.124\textwidth]{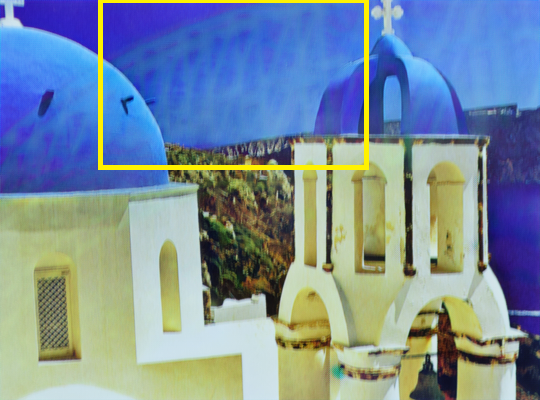}&\hspace{-4.2mm}
        \includegraphics[width=0.124\textwidth]{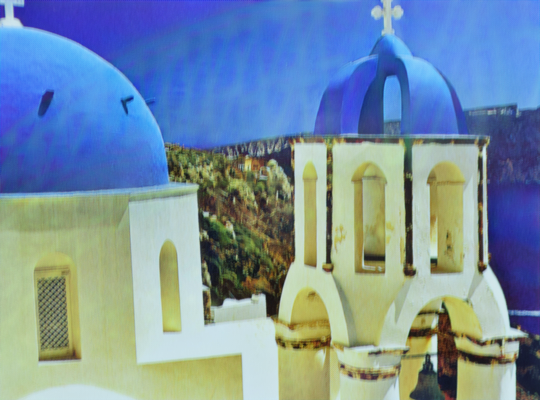}&\hspace{-4.2mm}
        \includegraphics[width=0.124\textwidth]{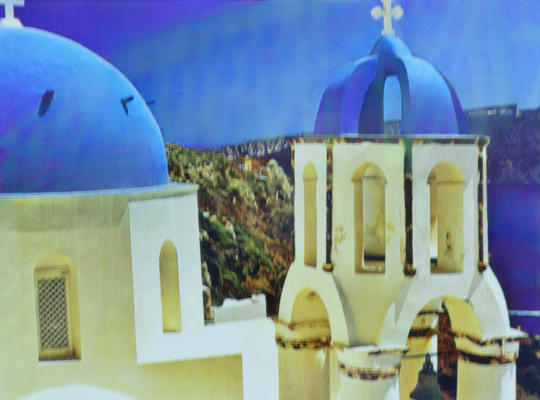}&\hspace{-4.2mm}
        \\
        Input&\hspace{-4.2mm}
        RAGNet~\cite{RAGNet}&\hspace{-4.2mm}
        RAGNet$_\mathit{OF}$&\hspace{-4.2mm}
        RAGNet$_\mathit{NI}$& \hspace{-4.2mm}
        \\
    \end{tabular}}
  \vspace{-3mm}
  \captionsetup{font={small}}
  \caption{Visual comparison of RAGNet based AdaNEC methods.}
  \label{fig:vis_ragnet}
  \vspace{-2mm}
\end{figure}

\subsection{Comparison with Backbone Methods}
\label{sec:exp_comparison}
As shown in \cref{tab:exp_main}, the peak signal-to-noise ratio (PSNR) and structural similarity (SSIM) indices of two AdaNEC methods are provided for comparing against each backbone method, which are denoted by the subscripts \textit{OF} and \textit{NI}, respectively.
We can see that, the output fusion (OF) manner generally can achieve an average PSNR gain of about 0.45$\sim$0.7 dB, and the performance on all datasets is boosted to some extent, showing the effectiveness of our AdaNEC$_\mathit{OF}$ method.
On the other hand, considering the computation amount and inference efficiency, the network interpolation (NI) manner also achieves considerable performance gain on most datasets.

It is worth noting that, both AdaNEC$_\mathit{OF}$ and AdaNEC$_\mathit{NI}$ achieve higher SSIM index than the backbone methods in all conditions, showing the promoted reflection removal ability.
For ERRNet~\cite{ERRNet} and RAGNet~\cite{ERRNet}, we also evaluate them on the \textit{Nature20} dataset, which is not considered by these methods.
As shown by the gray shaded regions in \cref{tab:exp_main}, the proposed method can generalize better to the unseen extra testing set.

As shown in \cref{fig:vis_errnet,fig:vis_ibcln,fig:vis_ragnet}, we also evaluate the proposed method qualitatively.
It can be seen that, when incorporated with our AdaNEC method from a domain generalization perspective, the backbone methods can be enhanced in terms of their reflection removal ability and output image quality.
Please zoom in for better observation, and refer to the supplementary material for more qualitative results.

\section{Ablation Study and Analysis}
\label{sec:ablation_study_analysis}

\subsection{Ablation Study}
\label{sec:ablation_study}
In this part, we conduct some ablation studies to show the contribution and influence of different parts of the proposed AdaNEC method.
Considering the training efficiency and relatively simple structure, the ablation studies are conducted on ERRNet~\cite{ERRNet}.

\begin{table}[t]
    \caption{Ablation studies on domain experts and IDE loss. The upper half shows the PSNR indices of domain experts trained with a single dataset, \ie, \textit{Syn$_\mathit{CEIL}$}, \textit{Real89}, and \textit{Unaligned}, where the results better than the backbone ERRNet~\cite{ERRNet} model are marked with \textbf{bold}. The lower half shows the performance of the model trained only with reflection removal losses (\ie, without $\mathcal{L}_\mathit{IDE}$).}
    \label{tab:abl_experts}
    \scalebox{0.82}{
    \begin{tabular}{lccccc}
        \toprule
         & \textit{Real20} & \textit{Wild} & \textit{Postcard} & \textit{Solid} & Average \\
         \cmidrule(lr){2-2} \cmidrule(lr){3-3} \cmidrule(lr){4-4} \cmidrule(lr){5-5} \cmidrule(lr){6-6}
         \multicolumn{1}{c}{\multirow{-2}{*}{Methods}} & PSNR$\uparrow$                & PSNR$\uparrow$                & PSNR$\uparrow$                & PSNR$\uparrow$                & PSNR$\uparrow$                \\ \midrule \midrule
        ERRNet~\cite{ERRNet}         & 22.08           & 25.13         & 22.76             & 24.62          & 23.79   \\ \midrule
        \textit{Syn$_\mathit{CEIL}$} & 20.22           & \textbf{25.31}& 22.45             & 24.39          & 23.51   \\
        \textit{Real89}              & \textbf{22.84}  & 24.58         & 20.24             & 24.52          & 22.66   \\
        \textit{Unaligned}           & 20.07           & 24.41         & \textbf{23.25}    & 24.56          & 23.80   \\ \midrule
        ERRNet$^\mathit{w\!/\!o}_ {\mathcal{L}_\mathit{IDE}}$           & 22.80           & 24.52         & 20.06    & 24.65          & 22.63   \\
        ERRNet$_\mathit{OF}$         & 22.80           & 25.26         & 23.08             & 25.26          & 24.24   \\ \bottomrule
    \end{tabular}}
\end{table}

\begin{figure}[t]
  \small
  \centering
    \begin{tabular}{cccccc}
      \hspace{-2mm}
      \includegraphics[width=0.23\linewidth]{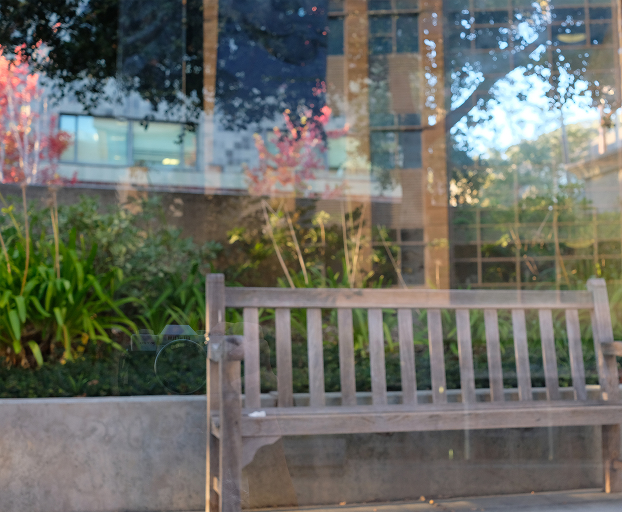}&\hspace{-4.2mm}
      \includegraphics[width=0.23\linewidth]{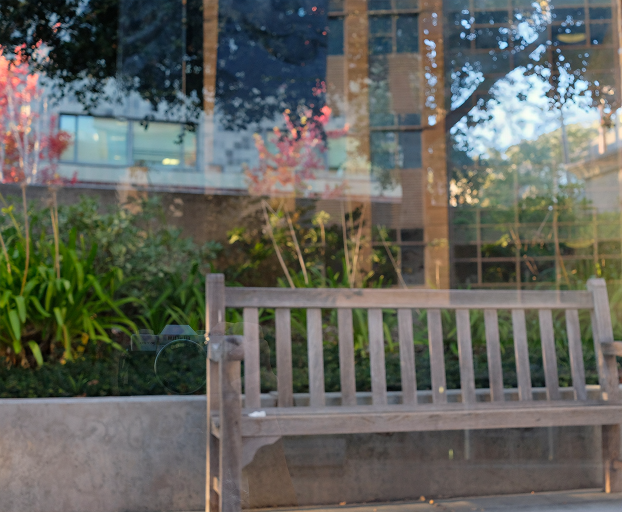}&\hspace{-4.2mm}
      \includegraphics[width=0.23\linewidth]{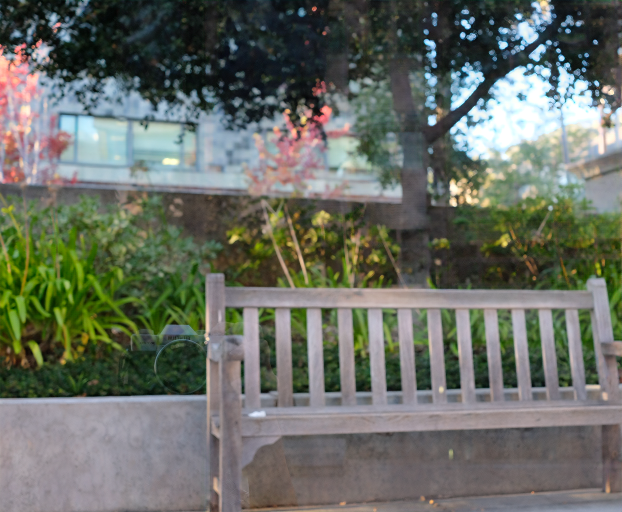}&\hspace{-4.2mm}
      \includegraphics[width=0.23\linewidth]{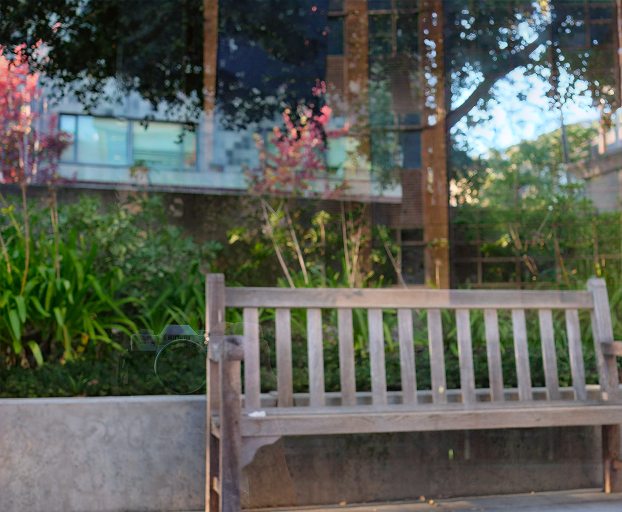}&\hspace{-4.2mm}
      \\
      \hspace{-2mm}
      Input&\hspace{-4.2mm}
      \textit{Syn$_\mathit{CEIL}$}&\hspace{-4.2mm}
      \textit{Real89}&\hspace{-4.2mm}
      \textit{Unaligned}& \hspace{-4.2mm}
      \\
      \hspace{-2mm}
      \includegraphics[width=0.23\linewidth]{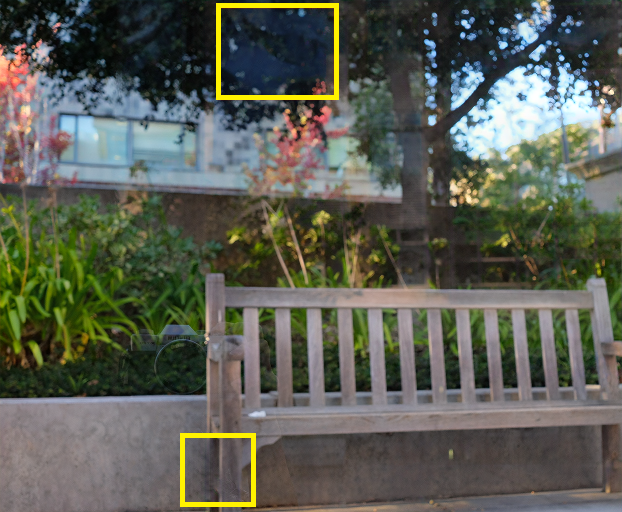}&\hspace{-4.2mm}
      \includegraphics[width=0.23\linewidth]{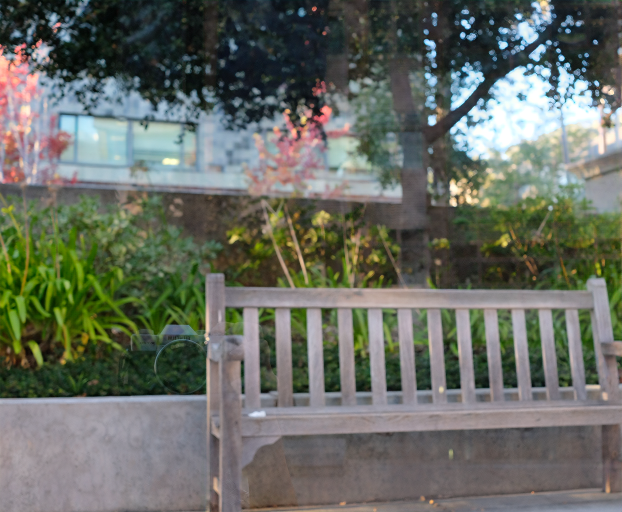}&\hspace{-4.2mm}
      \includegraphics[width=0.23\linewidth]{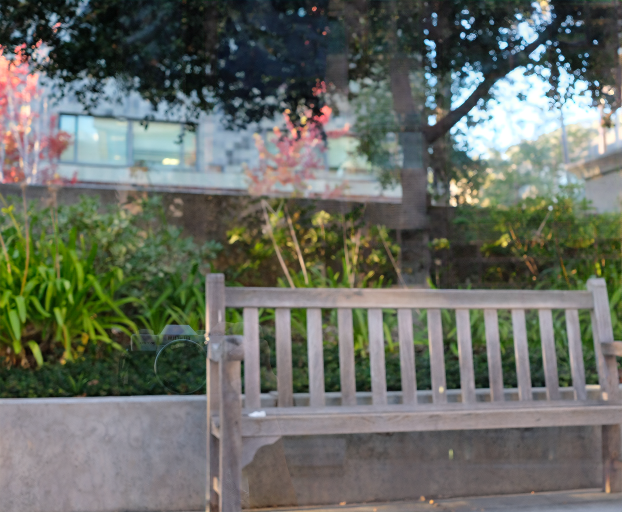}&\hspace{-4.2mm}
      \includegraphics[width=0.23\linewidth]{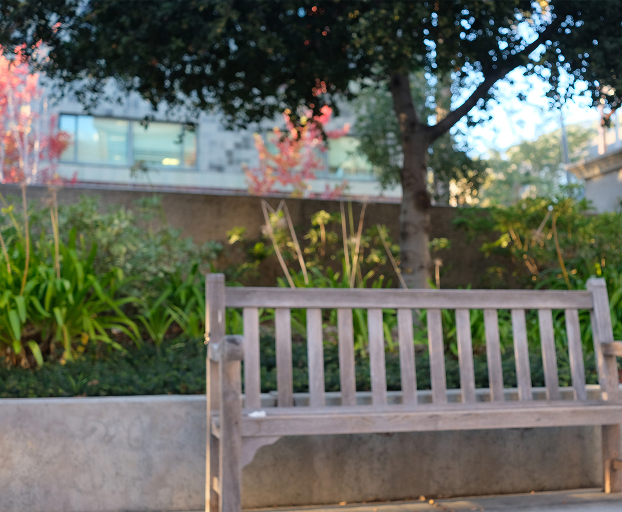}&\hspace{-4.2mm}
      \\
      \hspace{-2mm}
      ERRNet~\cite{ERRNet}&\hspace{-4.2mm}
      ERRNet$_\mathit{OF}$&\hspace{-4.2mm}
      ERRNet$_\mathit{NI}$& \hspace{-4.2mm}
      GT&\hspace{-4.2mm}
      \\

    \end{tabular}
  \vspace{-2mm}
  \captionsetup{font={small}}
  \caption{Visual comparison of ERRNet-based domain experts on \textit{Real20} dataset. Please zoom in for better observation.}
  \label{fig:abl_experts}
  \vspace{-2.5mm}
\end{figure}



\begin{table*}[t]
    \caption{Ablation studies on RTAW module. The indices better than the backbone model are highlighted with green and those worse than the backbone model are highlighted with red (the difference of PSNR $<$ 0.05dB and SSIM $<$ 0.004 has been ignored).}\vspace{-2mm}
    \label{tab:abl_rtaw}
    \centering
    \scalebox{0.9}{
    \begin{tabular}{lcccccccccc}
        \toprule
        \multicolumn{1}{c}{}                          & \multicolumn{2}{c}{\textit{Real20}}                                    & \multicolumn{2}{c}{\textit{Wild}}                                      & \multicolumn{2}{c}{\textit{Postcard}}                                  & \multicolumn{2}{c}{\textit{Solid}}                                     & \multicolumn{2}{c}{Average}                                   \\
        \cmidrule(lr){2-3} \cmidrule(lr){4-5} \cmidrule(lr){6-7} \cmidrule(lr){8-9} \cmidrule(lr){10-11}
        \multicolumn{1}{c}{\multirow{-2}{*}{Methods}} & PSNR$\uparrow$                & SSIM$\uparrow$                & PSNR$\uparrow$                & SSIM$\uparrow$                & PSNR$\uparrow$                & SSIM$\uparrow$                & PSNR$\uparrow$                & SSIM$\uparrow$                & PSNR$\uparrow$                & SSIM$\uparrow$                \\ \midrule \midrule
        ERRNet~\cite{ERRNet}                          & 22.08                         & 0.781                         & 25.13                         & 0.889                         & 22.76                         & 0.864                         & 24.62                         & 0.898                         & 23.79                         & 0.877                         \\ \midrule
        ERRNet$_\mathit{avg}$                         & \cellcolor[HTML]{FFDDDD}21.50 & \cellcolor[HTML]{FFDDDD}0.760 & \cellcolor[HTML]{66FF99}25.63 & \cellcolor[HTML]{66FF99}0.899 & \cellcolor[HTML]{66FF99}22.95 & \cellcolor[HTML]{66FF99}0.880 & \cellcolor[HTML]{66FF99}25.26 & \cellcolor[HTML]{66FF99}0.904 & \cellcolor[HTML]{66FF99}24.17 & \cellcolor[HTML]{66FF99}0.887 \\
        ERRNet$_\mathit{cl}$                          & \cellcolor[HTML]{FFDDDD}21.79 & \cellcolor[HTML]{FFDDDD}0.772 & 25.14                         & 0.888                         & \cellcolor[HTML]{66FF99}23.17 & \cellcolor[HTML]{66FF99}0.870 & \cellcolor[HTML]{66FF99}25.20 & 0.899                         & \cellcolor[HTML]{66FF99}24.20 & 0.880                         \\
        ERRNet$_\mathit{pl}$                          & 22.11                         & 0.782                         & \cellcolor[HTML]{FFDDDD}24.91 & \cellcolor[HTML]{FFDDDD}0.882 & \cellcolor[HTML]{66FF99}23.50 & \cellcolor[HTML]{66FF99}0.873 & \cellcolor[HTML]{66FF99}24.89 & \cellcolor[HTML]{FFDDDD}0.894 & \cellcolor[HTML]{66FF99}24.19 & 0.879                         \\ \midrule
        ERRNet$_\mathit{OF}$                          & \cellcolor[HTML]{66FF99}22.80 & \cellcolor[HTML]{66FF99}0.790 & \cellcolor[HTML]{66FF99}25.26 & 0.890                         & \cellcolor[HTML]{66FF99}23.08 & \cellcolor[HTML]{66FF99}0.874 & \cellcolor[HTML]{66FF99}25.26 & \cellcolor[HTML]{66FF99}0.904 & \cellcolor[HTML]{66FF99}24.24 & \cellcolor[HTML]{66FF99}0.885 \\ \bottomrule
    \end{tabular}}
\end{table*}

\vspace{0.5em}
\noindent\textbf{Performance of domain experts.}
As shown in \cref{tab:abl_experts}, when trained with a single dataset, the performance of the domain experts may be enhanced on some datasets (\eg, generally the performance on \textit{Real20} will be better when trained with \textit{Real89}, since they are the training/testing set from the same dataset), and the final output fusion (OF) results can approximate these enhanced performance.
For \textit{Solid}, all these domain experts perform slightly worse than the backbone model.
However, a decent performance gain is obtained when incorporating with AdaNEC, which shows the generalization ability of the proposed method.
We also show the visual results in \cref{fig:abl_experts}, and the AdaNEC-boosted models can generate better reflection removal results.

\vspace{0.5em}
\noindent\textbf{In-domain Expert (IDE) Loss.}
As illustrated in \cref{sec:method_loss}, the IDE loss is introduced from a straightforward idea.
On the one hand, it explicitly requires that the reflection removal should rely more on the in-domain expert, which accords with the idea of determining the weights according to the expertise level in domain generalization.
On the other hand, if the RTAW module is trained only with the reflection removal losses, it will keep pursuing extreme weights for some domain experts, resulting in large and unstable values in the RTAW outputs (\ie, $\mathit{v}_\mathit{i}$).
When the IDE loss is applied, the weight of other domain experts will be constrained to a reasonable range, and we empirically find that the training of RTAW module will become more stable.

As shown in \cref{tab:abl_experts}, when the IDE loss is discarded (denoted by ERRNet$^\mathit{w\!/\!o}_{\mathcal{L}_\mathit{IDE}}$), the training of RTAW is usually unstable and may result in different order of magnitude of the expertise level ($\mathit{v}_\mathit{i}$).
Therefore, the final result heavily relies on a specific domain expert.
We can see that the performance of ERRNet$^\mathit{w\!/\!o}_{\mathcal{L}_\mathit{IDE}}$ is very close to the domain expert trained on \textit{Real89} dataset, indicating the failure of training.

\vspace{0.5em}
\noindent\textbf{RTAW Module.}
We also conduct ablation studies on the structure of RTAW module, and three variants are compared in the following.
1)~ERRNet$_\mathit{avg}$: A plain average operation, which is the simplest way to perform both output fusion and network interpolation.
2)~ERRNet$_\mathit{cl}$: A classification model has been deployed in \cref{sec:method_analysis}, which shows the ability to distinguish the source of a reflection-contaminated image. Hence, it is reasonable to take such an architecture as an RTAW module variant.
3)~ERRNet$_\mathit{pl}$: In this case, $\mathit{N}$ classification models are deployed in parallel, where each model predicts the expertise level for a specific domain expert.

As shown in \cref{tab:abl_rtaw}, the generalization ability is not guaranteed when a plain average operation is taken, resulting in performance degradation of ERRNet$_\mathit{avg}$ on the \textit{Real20} dataset.
ERRNet$_\mathit{cl}$ and ERRNet$_\mathit{pl}$ also face the problem of performance degradation on some datasets.
On the contrary, the RTAW architecture with cross-domain attention module (\ie, ERRNet$_\mathit{OF}$) can generalize better to all target domains, showing the superiority of the proposed RTAW design.

\subsection{Limitations and Future Work}
\label{sec:limitations}
As shown in \cref{tab:exp_main}, when using the IBCLN~\cite{IBCLN} backbone, the network interpolation (NI) manner does not work well and brings minor performance gain. This phenomenon should be attributed to the loop structure, which usually leads to accumulated errors as observed by Li \etal~\cite{RAGNet}, and the interpolated parameters may exacerbate this effect.
Besides, as shown in \cref{tab:abl_experts}, although the proposed method can boost the reflection removal performance, it still does not reach the full potential of all domain experts (\eg, the qualitative performance of ERRNet$_\mathit{OF}$ on \textit{Postcard} is worse than the domain expert trained on \textit{Unaligned}).

As a remedy, in the future, we will explore the specialized architecture to better exploit the power of domain generalization methods.
It is worth noting that the proposed cross-domain attention module (CDAM) can be naturally expanded to a multi-head version like the self-attention layer in Transformers~\cite{transformer}, which can be used as a component in such an architecture for converting various amount of source domains into a fixed number of features.

\subsection{Ethical Discussions}
\label{sec:ethical}
The reflection removal methods can promote the image quality and benefit subsequent tasks like autonomous driving, the datasets are either from postcards, solid objects, natural scenes or allowed for academic usage, therefore there are no ethical issues or negative societal impacts.

\section{Conclusion}
\label{sec:conclusion}
In this paper, we analyzed the domain gaps between reflection-contaminated images, and propose to enhance the learning-based single-image reflection removal (SIRR) methods from a domain generalization perspective.
Given domain experts of various reflection types, a reflection type-aware weight (RTAW) module is introduced to predict the expert-wise weights, which are used for adaptive network combination (AdaNEC).
An in-domain expert loss is proposed for training the RTAW module.
Experimental results show that the proposed AdaNEC can bring appealing performance gain on different state-of-the-art SIRR networks.

\clearpage
{\small
\bibliographystyle{ieee_fullname}
\bibliography{egbib}
}

\end{document}